\def\year{2019}\relax
\documentclass[letterpaper]{article} 
\usepackage{aaai19}  
\usepackage{times}  
\usepackage{helvet}  
\usepackage{courier}  
\usepackage{url}  
\usepackage{graphicx}  
\usepackage{amsmath}
\begin{document}
%
\title{Explainable Failure Predictions with RNN Classifiers based on Time Series Data}
\author{Ioana Giurgiu\\
IBM Research - Zurich\\
igi@zurich.ibm.com
\And Anika Schumann\\
IBM Research - Zurich\\
ikh@zurich.ibm.com}
\maketitle
\begin{abstract} 

Given key performance indicators collected with fine granularity as time series, our aim is to predict and explain failures in storage environments. Although explainable predictive modeling based on spiky telemetry data is key in many domains, current approaches cannot tackle this problem. Deep learning methods suitable for sequence modeling and learning temporal dependencies, such as RNNs, are effective, but opaque from an explainability perspective. Our approach first extracts the anomalous spikes from time series as events and then builds an RNN classifier with attention mechanisms to embed the irregularity and frequency of these events. A preliminary evaluation on real world storage environments shows that our approach can predict failures within a 3-day prediction window with comparable accuracy as traditional RNN-based classifiers. At the same time it can explain the predictions by returning the key anomalous events which led to those failure predictions.


\end{abstract}

\section{Introduction}

Explainable predictive modeling based on telemetry data is key in many domains, from healthcare to IT and industries, and it is particularly challenging when concerned with critical incidents. In IT environments, these incidents represent failures of devices or components and are typically very rare ($<$ 2-3\% of all incidents). Even being able to predict a small fraction of them significantly increases availability, generates savings and avoids labor costs, as opposed to taking the current approach where maintenance and repair operations are performed reactively, after the critical incidents occur.

As a result, up until recently the primary concern from an AI perspective was to build models that can predict such failures ahead of time as accurately as possible. This has led to a transition from traditional ML approaches, such as random forests and gradient boosted machines, to deep-learning methods because of their superior performance. In particular, RNNs are effective for use cases that benefit from  sequence modeling and learning temporal dependencies. However, in the last few years it has become clear that such predictive models are only applicable in practice if they provide some degree of usable intelligence~\cite{crafting,pneumonia}. That implies they need to be able to learn from prior data, generalize well and extract the explanatory factors of the data~\cite{representation}. Therefore, the current objective is to build accurate AI models that at the same time provide explanations intelligible to domain experts. Numerous approaches have been proposed~\cite{lime,layerwise,decisionsets,anchors,pairwise,additive}, either providing post-hoc explainability (i.e., agnostic to the underlying black-box model) or ante-hoc explainability (i.e., incorporating explanations in the black-box model itself). What they have in common is their application primarily on images and text. This is most probably due to two reasons: 1) explanations around text or images are easier to comprehend by humans (e.g., an explanation highlighting a guitar in an image makes sense for a classifier that is supposed to determine the presence or absence of a guitar, and provides confidence in the underlying classifier), and 2) there is no time component involved.

In this paper, we are concerned with explaining predicted failures in storage environments, based on key performance indicators (KPIs) collected with fine granularity as time series. Given the temporal component, using RNNs as the underlying classifier is a natural choice. While using post-hoc explainable models is attractive, since it does not require changes to the black-box model itself, to the best of our knowledge there are no approaches built specifically for temporal data. We show that applying LIME~\cite{lime} to our time series KPIs floods the human expert with a vast number of imprecise explanations in the shape of highlighted portions of the series. 

Therefore, we consider incorporating explainability directly into the classifier. Based on the observations that our KPIs are spiky rather than exhibiting increasing or decreasing trends (Fig.~\ref{fig:spiky}), and that the progression and accumulation of spikes over time can lead to critical incidents, we are inspired by the medical domain, where RNN-based approaches for diagnosis have modeled the temporal progression of an illness as event series with decaying factors~\cite{disease}. Therefore, we transform our time series into series of clustered anomalous events, where these events are defined as $KPI_{val}^{t} > threshold$ ($KPI_{val}^{t}$ is the value of a KPI at time $t$). Such clusters exist, because anomalous events have a tendency to co-occur within well-defined time intervals. Moreover, they appear at random points in time and incur a bursty behavior -- long periods with no or few events, followed by shorter periods with multiple events. Traditional RNNs are oblivious to the time interval between two clusters. Recent efforts use the information about these time intervals to compute the hidden states of the recurrent unit~\cite{rnnmultivariate,retain,irregularity}, but do not consider that the health status of a device and the time between events are correlated. Intuitively, the more frequent and recent the anomalous events, the more impact they will have on future critical incidents.

In short, our approach follows multiple steps: 1) clusters (windows) of anomalous events are detected optimally via Ckmeans.1d.dp~\cite{ckmeans}; 2) unique anomalous events are embedded in a continuous vector space; 3) for each event in a cluster, context information is aggregated using the attention mechanism proposed in~\cite{attention}; 4) for each event, we build a temporal progression function that quantifies how much of an impact the event has on the prediction objective, depending on its type and when it occurred; 5) using the context information and the progression function, each window is represented as a weighted sum of embeddings of its events; and 6) the window representations are used in an LSTM network to predict failures within a predefined interval. We note that while multiple anomalous events of the same type can occur in a window (Fig.~\ref{fig:windows}), no progression is assumed within the same window.

We conducted a preliminary evaluation on KPIs collected from 130+ storage environments over 14 days in May and June 2018, respectively, with 5 minute granularity. Based on threshold rules defined by domain experts, we extract 266081 anomalous events and use them to predict storage failures within 3 days after each 14-day interval. Initial results show that our approach can achieve comparable accuracy with traditional RNN-based models, while at the same time it provides useful insights into its prediction decisions.

\section{Motivation}

Our approach is driven by the characteristics of our data.

\begin{figure}
  \includegraphics[width=\linewidth]{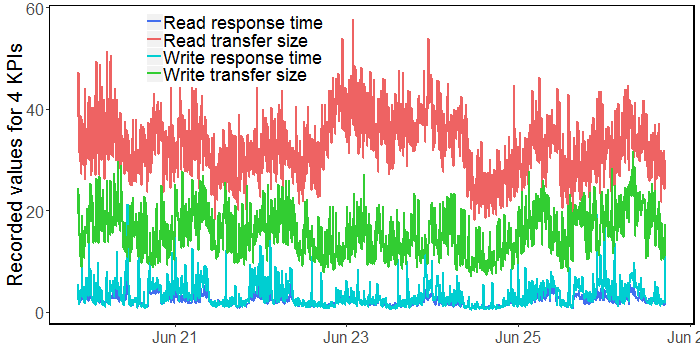}
  \caption{Time series of Read/Write response time and Read/Write transfer size collected over 10 days for a device. }
  \label{fig:spiky}
\end{figure}

\textit{Spiky nature of KPIs} -- Although there is an expectation that KPIs should exhibit either an increasing or decreasing trend prior to a critical event, it is very common for them to actually be spiky in nature (with their values at times exceeding pre-defined thresholds), but maintain relatively constant means through time. In Fig.~\ref{fig:spiky}, we show such an example, namely the time series collected for 4 KPIs over a period of 10 days for one storage device that fails 2 days later. To confirm the spikeness of our data beyond visual inspection, we apply changepoint detection via causal Bayesian inference~\cite{changepoint}. The underlying premise of changepoint detection is that when a KPI has impact over a future critical incident, there should be a significant shift at a timestamp before the incident occurs. Even more important, this shift should be permanent. Shortly, for a time series $S_i=(s_1, s_2,\cdots,s_p)$ of KPI $i$, if there exists a timestamp $t<p$ when a significant change in $S_i$ occurs (e.g., values start increasing), then $S_i$ potentially has impact over a future critical event. To verify whether the change is permanent, we check if the difference between the time series of the KPI and the corresponding time series in the absence of the change at time $t$ (i.e., synthetic series) is significant. To generate the synthetic series, we compute the posterior distribution $p(s_{t+1:p} \mid s_{1:t}, x_{1:p})$, where $s_{1:t}$ is the series in the pre-change period and $x_{1:p}$ is the control time series generated from healthy devices. Finally, the sensor metric has impact if the probability distributions of the actual time series after the detected changepoint and the synthetic series are significantly different (i.e., p $<$ 0.05). With all our KPIs, the probability distributions are not different (p $\gg$ 0.05).

\begin{figure}
  \includegraphics[width=\linewidth]{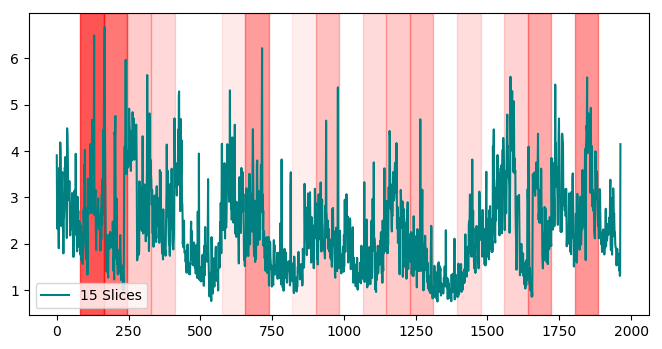}
  \caption{Highlighted slices for Read response time series over 10 days, extracted with LIME (15 slices, 82 points per slice). Darker red indicates higher contribution to prediction.}
  \label{fig:lime}
\end{figure}

\begin{figure}
  \includegraphics[width=\linewidth]{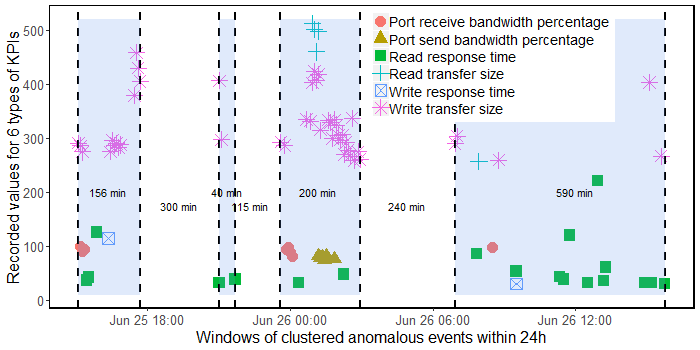}
  \caption{Anomalous events clustered in 4 highlighted time windows of variable length during 24h for a storage device.}
  \label{fig:windows}
\end{figure}

\textit{LIME for time series} -- Even considering the spiky nature of the KPIs, it is still attractive to apply agnostic models, such as LIME~\cite{lime}, to understand how various KPIs contribute to a prediction. Therefore, we use a binary classifier to tackle the prediction problem introduced previously -- taking as input KPI time series collected over 14 days to predict whether a failure will occur within the next 3 days. Then, we apply a parametric time series implementation of LIME~\cite{limets} to understand which KPIs and portions of their corresponding series contribute to individual predictions. Fig.~\ref{fig:lime} highlights the portions for read response time from Fig.~\ref{fig:spiky} that contributed to the failure incident recorded 2 days later. We note the following limitations: 1) the quality of the explanation highly depends on the number of slices given as an input parameter and it involves significant trial and error; 2) the highlighted portions in the series have a fixed length (i.e., size of the slice); 3) the lower the number of slices, the less discriminative a series portion becomes; 4) more slices result in a vast number of explanations per KPI that are difficult to follow even by a domain expert; 5) there is no temporal consideration -- in the example, the highest contribution is attributed to the highest jumps in the read response time, which happen to occur in the first day of the series (12 days before the failure). This is completely opposite to how a system behaves, namely the more recent the spikes, the more impact they will have on future critical incidents.

\textit{Events co-occurence} -- Driven by the spiky behavior of the KPIs and the limitations of applying LIME, we consider using the spikes as anomalous events rather than the entire time series. Spikes can be captured either by basic thresholding (e.g., read response time $\geq$ 200 ms/op) or by detecting them as anomalies using AI techniques. While numerous anomaly detection algorithms have been proposed in recent years~\cite{nemo,anomaly1,anomaly2}, the unsupervized phase is always followed by a semi-supervized phase, namely the anomaly validation. Typically validating anomalies has to be done with a human in the loop, even if the manual validation effort would only involve a small sample of the detected anomalies. For the purpose of this paper, we revert to using expert-defined thresholds to capture the spikes. As a result, we uncover an interesting pattern: these spikes have a tendency to co-occur within relatively compact time windows, separated by windows where no spikes appear. Fig.~\ref{fig:windows} highlights 4 windows of clustered spikes captured within 24h for a storage device. All windows are of variable length and contain an arbitrary number of spikes recorded pertaining to 6 KPIs. For all other KPIs, no spikes exceeding pre-defined thresholds were recorded. The challenge is to detect these windows. We detail our method next.

\section{Method}

In this section, we describe our proposed prediction model. First, we introduce the problem setup. Second, we describe how we detect the windows of anomalous events. Then, we detail our RNN-based approach, and finally we explain how we provide explanations based on analyzing weights associated to anomalous events.

\subsection{Problem Setup}
Our dataset is a collection $D$ of $d$ storage devices. For each device, a set $KPI$ of $M$ metrics are collected as time series at regular time intervals over a period $t$. Across all $M$ time series pertaining to a device $D_{i}$, we collect the set of $l$ anomalous events $E_{D_{i}}=\{{e_{1D_{i}}, ..., e_{lD_{i}}}\}$. Each anomalous event refers to one KPI and is registered when a pre-defined threshold is exceeded, such that $KPI_{j} > th_{KPI_{j}}$. Next, we cluster the events for each device $D_{i}$ into a set of windows $W_{D_{i}}$ = $\{W_{1D_{i}}, ..., W_{pD_{i}}\}$, where $p < l$, ordered chronologically. Window $W_{rD_{i}}$ is a 2-dimensional vector $W_{rD_{i}}$ = $\{\beta_{rD_{i}}, t_{rD_{i}}\}$ of unordered anomalous events $\beta_{rD_{i}} = \{e_{1D_{i}},... e_{|\beta_{rD_{i}}|D_{i}}\}$ and corresponding timestamps. We denote with $\mathbf{E}$ the vocabulary of events and its cardinality with $|\mathbf{E}|$.

\subsection{Detecting Windows of Anomalous Events}
In the first stage, we detect all windows of anomalous events for a device, $W$ = $\{W_{1}, ..., W_{p}\}$, $p < l$, within the time interval [$0, t$]. Since our problem is one-dimensional (1-D), we use an optimal k-means clustering algorithm with dynamic programming, called Ckmeans.1d.dp~\cite{ckmeans}. We essentially assign anomalous events of the input 1-D array $E$ into $k$ clusters so that the sum of squares of inner-cluster distances from each event to its corresponding cluster mean is minimized. The beauty of Ckmeans.1d.dp is that it can estimate $k$ optimally on its own, using Bayesian information criterion. An example of detected windows is shown in Fig.~\ref{fig:windows}. As seen, all windows are of arbitrary length and clearly separated one from the other.

\subsection{Representing Windows of Anomalous Events}
Next, the objective is to predict future critical incidents -- particularly failures -- for a device $D_{i}$, within a window [$t$, $t + T$], by using the corresponding windows of anomalous events, $W_{1}, ..., W_{p}$. An essential aspect is how we represent each window $W_{r}$, $r < p$, as a vector. A simple approach is to embed every event in $\mathbf{E}$ in a continuous vector space, such that the $s$-dimensional vector $v_{e}$ $\in$ $R^{s}$. Then, the vector representation of a window $W_{r}$, $w_{r}$, containing $N$ events $\beta_{r} = \{e_{1}, ..., e_{N}\}$ is the sum of embeddings of events occuring within the window, namely
\begin{equation}
w_{r} = \sum_{n=1}^{N} x_{e_{n}} v_{e_{n}}
\end{equation}
\noindent where $x_{e_{n}}$ represents how many times event $e_{n}$ occurred within window $W_{r}$. The problem with this approach is that the impact of an event is proportional to its frequency. If all events occur only once, they contribute equally to the window representation. Instead, the window representation should also depend on when an event occurs and what it is. Given that we treat each window as a set of unordered events (since we cannot assume causality in the events sequence), we consider using attention mechanisms
~\cite{attention}, which have been successfully applied in language translation. We define the context vector $cv_{n}$ for an event $e_{n}$ as 
\begin{equation}
cv_{e_{n}} = \sum_{x=1}^{N} {\alpha}_{nx} v_{e_{x}}
\end{equation}
\noindent where the attention value is $\alpha_{n1}, ..., \alpha_{nN} = softmax([\frac{q_{n}k_{1}}{\sqrt{a}}, ..., \frac{q_{n}k_{N}}{\sqrt{a}}])$, as defined in~\cite{disease}. $q_{n}$ is the query vector for $e_{n}$, $k_{n}$ is the key vector for the same event and $a$ is their dimension. 

Now that the context vector for an event $e_{n}$ in a window is defined, we need to quantify each event's contribution to the prediction representation in [$t$, $t + T$]. As already stated, the contribution of each event depends on when the event occurred. The further in the past, the smaller the contribution. Therefore, we define the contribution as follows:
\begin{equation}
I(c_{e{_n}}, \Delta) = S(\theta_{e_{n}} - \sigma_{e_{n}}\Delta) \in [0,1]
\end{equation}
\noindent where S is the sigmod function, $\theta_{e_{n}}$ is the initial contribution of $e_{n}$ to the prediction, $\sigma_{e_{n}}$ is the progression of this contribution function of time and $\Delta$ is the time elapsed from window $W_{j}$ to the end of prediction window, namely $\Delta = t + T - \tau_{W_{j}}$. 

Finally, we rewrite the vector representation of a window $w_{r}$ as follows:
\begin{equation}
w_{r} = \sum_{n=1}^{N} x_{e_{n}} I(c_{e_{n}}, \Delta) cv_{e_{n}}
\end{equation}

\subsection{Explanaible Predictions with LSTMs}
We use the window representations, $w_{1}, ..., w_{p}$, as inputs to the LSTM to predict a label $y$ $\in$ \{0,1\}, which represents whether a critical event will occur or not in the [$t, t+T$] interval. Note that our problem reduces to binary classification and the predicted event does not belong to the vocabulary $\mathbf{E}$, such that:
\begin{equation}
\hat{y} = \sigma(W h_{p} + b)
\end{equation}
\noindent where $h_{p}$ is the hidden state output at step $p$ of the LSTM, $W$ is the weight matrix, $b$ is the bias vector of the output function and $\hat{y}$ is the predicted label probability distribution.

We explain predictions by quantifying how much each anomalous event within a window $W_{r}$ contributed to the decision. For each embedding $v_{e_{n}}$ of event $e_{n}$, there are associated weights defining that contribution. More specifically,
\begin{equation}
\begin{split}
w_{r} = \sum_{n=1}^{N} x_{e_{n}} I(c_{e_{n}}, \Delta) \sum_{x=1}^{N} {\alpha}_{nx} v_{e_{x}} \\
          = \sum_{x=1}^{N}(\sum_{n=1}^{N} x_{e_{n}} I(c_{e_{n}}, \Delta)\alpha_{nx} v_{e_{x}})
\end{split}
\end{equation}
\noindent where $\sum_{n=1}^{N} I(c_{e_{n}}, \Delta)\alpha_{nx}$ is the contribution of each event embedding and $ x_{e_{n}}$ is the frequency of occurrence of each event. 

\section{Preliminary Evaluation}
\subsection{Data and Setup}
\textit{KPIs and Events} -- We collect KPIs as time series for 130+ storage environments over 14 days in May and June 2018, respectively, with 5 minute granularity. As storage environments are typically complex, the KPIs are collected across their entire hierarchy (e.g., nodes, ports, volumes, RAID arrays, disks). More specifically, each storage device is mapped to multiple nodes (2-4 on average) connected to tens or hundreds of hosts, through a varying number of ports (a power of 2 in the range [4,32], each with 8 or 16 GB/s speed), depending on the fabric architecture. For each environment, we collect all the anomalous events registered for both 14 days intervals, according to the threshold rules defined by domain experts. In total, there are 266081 such events, distributed across 15 KPI rules, as shown in Table~\ref{datadesc}.

\textit{Critical incidents} -- Additionally, we collect all the incidents registered within 3 days of the end of each time interval, as these represent our prediction goal. These incidents capture a variety of problems, such as "running out of space", "device health is below standard", "drive has excessive errors interfering with the hardware", "software level is below recommended version", "battery is at end of life" or "drive not responding to commands". They are classified based on severity in \textit{informational} (87\%), \textit{warning} (9\%) and \textit{error} (4\%). We focus on \textit{error} incidents and keep only those regarding devices or drives and containing the phrase "the device or drive is likely to fail soon". From hundreds of incidents collected, these are less than 2\% and refer to 3\% of the devices. This implies a 1:32 ratio between the minority (i.e., failure) and majority classes.

\textit{Design and Metrics} -- We compare our model with two other binary classifiers, namely random forest (RF) and traditional LSTMs. The RF model, implemented in R, is optimized to use the optimal number of trees, oversampling ratio of the minority class and the number of samples at leaf nodes. In addition, we assign double weight to the minority class to penalize miss-classified failures more. The LSTM model is a traditional network, where the sum of the event embeddings are used as window representations (Equation 1) and fed as inputs. It is implemented in Keras and uses a batch size of 50, a sequence length of 15 and a learning rate of 0.1. We use the same LSTM parameters for our model. For both datasets, we randomly sample 80\% for training and the rest of 20\% for testing and validation. We report precision and F1-score values for the minority class, as well as the balanced accuracy across both classes.

\begin{table}
    \begin{tabular}{| l | l |l|}
\hline
\textbf{KPI} & \textbf{Threshold} & \textbf{\#} \\
\hline
Disk utilization & 50 \% & 4448 \\
\hline
Invalid transmission word rate & 0.7 cnt/s &  4331 \\
\hline
Peak backend write resp. time & 10s & 238 \\
 \hline
Port receive bandwidth & 75\% & 4314 \\
 \hline
Port send bandwidth & 75\% & 332 \\
 \hline
Port send delay I/O & 20\% & 55170 \\
 \hline
Port to local node send queue time & 0.5ms/op & 720\\
\hline
Port to local node send resp. time & 0.75ms/op &20666\\
 \hline
Read response time & 30ms/op &  49010 \\
 \hline
Read transfer size & 64KB/op & 79100 \\
\hline
System CPU core utilization & 70\% & 706 \\
\hline
Write-cache delay & 3\% & 112 \\
\hline
Write response time & 30ms/op & 52424 \\
\hline
Write transfer size & 256KB/op & 44473 \\
\hline
Zero buffer credit & 20\% & 37 \\
\hline
    \end{tabular}
\caption{Anomalous events distributed across 15 KPI rules.}  
\label{datadesc}
\end{table}

\subsection{Windows Size}
\begin{figure}[h]
\begin{minipage}[t]{0.45\linewidth}
    \includegraphics[width=\linewidth]{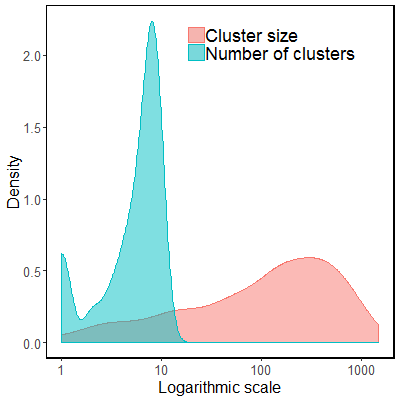}
    \caption{Distribution of cluster sizes and number of clusters per device.}
    \label{fig:clusters}
\end{minipage}%
    \hfill%
\begin{minipage}[t]{0.45\linewidth}
    \includegraphics[width=\linewidth]{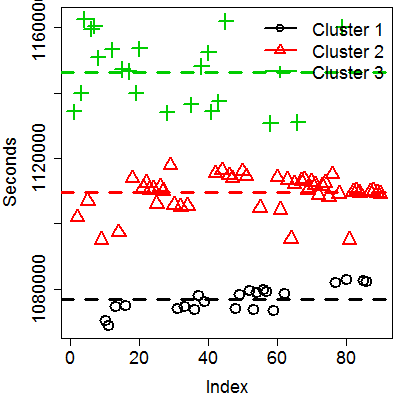}
    \caption{Example of 3 clusters for a device with 90 anomalous events. }
    \label{fig:example}
\end{minipage} 
\end{figure}

We analyze the number of windows obtained when clustering anomalous events per each device, as well as their size (i.e., the number of events within a window). The distributions are shown in Fig.~\ref{fig:clusters}. Typically, the number of clusters varies from 1 to 10 (rarely to 15) and the size can go up to 1500 events. Single clusters per device are very rare and appear only when there are few events recorded in a short time span. These are specific to devices that have fewer ports and nodes, and are connected to less hosts. For devices that have very complex environments (e.g., hundreds of servers connected and hundreds or thousands of volumes), we record thousands of anomalous events. In this case, both the number of the clusters and their size increases, which explains the shapes of the distributions. Fig.~\ref{fig:example} shows the 3 clusters obtained with the Ckmeans.1d.dp algorithm for a device with 90 anomalous events recorded over 24h. The clusters are of sizes 22, 47 and 21, respectively.

\subsection{Prediction Accuracy}

Table~\ref{accuracy} shows the accuracy and F1-score values for the minority class, as well as the balanced accuracy across both minority and majority classes, obtained with the RF, LSTM and LSTM with attention models. On the one hand, RF is outpeformed by both neural network models, which is to be expected considering that RFs are not built to specifically deal with temporal data. On the other hand, the gains in accuracy obtained with both LSTM models are relatively modest. We attribute this to: 1) the challenge of the prediction problem, since the data set is highly imbalanced (1:32 ratio between minority and majority classes); 2) the lack of hyper-parameter optimization for both LSTM implementations.

However, our goal was to build a model that does not sacrifice accuracy in favor of explainability. As seen, the LSTM with attention approach is slightly better than the traditional LSTM, while at the same time being able to provide prediction interpretability, as shown next.

\begin{table}
\centering
    \begin{tabular}{| l | l | l | l |}
\hline
\textbf{Model} & \textbf{Prec.1} & \textbf{F1.1} & \textbf{bACC} \\
\hline
RF & 0.24 & 0.19 & 0.55 \\
\hline
LSTM & 0.28 &   0.23 & 0.59\\
\hline
LSTM with attention & \textbf{0.29} & \textbf{0.24} & \textbf{0.60} \\
 \hline
    \end{tabular}
\caption{Precision and F1-score on the minority class, and balanced accuracy for RF, LSTM and LSTM with attention.}  
\label{accuracy}
\end{table}

\subsection{Explainability}
The model provides explainability by associating weights with each anomalous event as a function of its frequency within each window and the contribution of its embedding (Equation 6). We show how our approach works for 3 storage devices for which our predictions indeed occur within the prediction window. The first device is predicted to fail with a probability of 0.87, while the second and third are predicted to fail with a probability of 0.23 and 0.31, respectively (i.e., 0.77 and 0.69 probabilities not to fail).

\begin{table*}
\centering
    \begin{tabular}{| l | l | l | l | l | l | l |}
\hline
\textbf{Window} & \textbf{Start timestamp} & \textbf{Duration} & \textbf{\# Events} &\textbf{Event} & \textbf{Frequency} & \textbf{Contribution} \\
\hline
1 & Day 1 22:58 & 115 min & 11& \shortstack[l]{Read response time \\Read transfer size \\Write transfer size} & \shortstack{1 \\ 5 \\ 5} & \shortstack{0.00\\0.00\\0.00} \\
\hline
2 & Day 2 6:04 &105 min & 1& Read response time & 1 & 0.00 \\
\hline
3 & Day 2 20:19&390 min&8& \shortstack[l]{Read response time \\Read transfer size \\Write transfer size} & \shortstack{1 \\ 3 \\ 4} & \shortstack{$<$0.01\\$<$0.01\\0.04}\\
 \hline
4 & Day 3 19:59& 195 min &6& \shortstack[l]{Read transfer size \\Write transfer size} & \shortstack{3\\3} & \shortstack{$<$0.01\\0.06} \\
\hline
5 & Day 4 23:00 & 70 min&3& \shortstack[l]{Read transfer size\\Write transfer size}& \shortstack{1\\2}&\shortstack{$<$0.01\\0.06}\\
\hline
6 & Day 5 6:15 & 120 min &2& Read response time & 2& 0.015\\
\hline
7 & Day 5 22:55 & 20min &2& Read response time & 2 & 0.02 \\
\hline
8 & Day 6 22:56 & 20 min &1& Read transfer size & 1 & $<$0.01\\
\hline
9 & Day 7 23:01 & 15 min &2& Read transfer size & 2 & 0.01\\
\hline
10 & Day 8 6:02 & 125 min &3& Disk utilization & 3 & 0.00 \\
\hline
11 & Day 8 22:57 & 20 min &9& \shortstack[l]{Read transfer size \\Write transfer size}& \shortstack{5\\4} & \shortstack{0.05\\0.16} \\
\hline
12 & Day 9 23:12 & 65 min &3& Read response time & 3 & 0.06 \\
\hline
13 & Day 11 20:28 & 205 min &4& Write response time & 4 & 0.18 \\
\hline
14& Day 13 4:08 & 35 min &6& \shortstack[l]{Read response time\\Write response time} & \shortstack{4\\2} & \shortstack{0.1\\0.34} \\
\hline
15 & Day 14 22:59 & 15 min &8& \shortstack[l]{Read response time \\Peak backend write response time\\Write response time} & \shortstack{3\\2\\3} &  \shortstack{0.12\\0.8\\0.63} \\
\hline
    \end{tabular}
\caption{Weights associated with 69 anomalous events clustered in 15 windows for a storage device predicted to fail.}  
\label{fail}
\end{table*}

\textit{Prediction: Fail} -- For this device, there are 69 anomalous events recorded during 14 days in June. These events are clustered in 15 windows as shown in Table~\ref{fail}. For each window, we show the number of events distributed by type and the weights associated. As seen, most assigned weights are either 0 or $<$ 0.01 as they do not contribute to the prediction. Such sparsity in the weights distribution is useful for domain experts interpreting a prediction, since the model focuses on events with non-negligible predictive power. For instance, our approach strongly prefers the events occuring in the last 3 windows, and more specifically, those related to write response times thresholds being exceeded (i.e., write response time and peak backend write response time). While higher than expected write response times are already signaling performance decline, together with peak backend write response times (i.e., the longest time for a back-end storage resource to respond to a write operation by a node) they would indicate an impeding critical incident, such as a failure. Our model learns such associations between anomalous events. In addition, it learns that read response time events contribute significantly less to the failure (i.e., 0.04 in window 15 as opposed to 0.4 and 0.21 for peak backend write response time and write response time, respectively) and their weights decrease the more distant they are to the prediction window.

\textit{Prediction: No fail / Example 1} -- Table~\ref{nofail} shows the weights associated to 22 anomalous events clustered in 7 windows for a device that remains healthy throughout the 3-day prediction window. We note the following. First, the model learns that events such as disk utilization threshold exceeded do not lead to critical incidents (i.e., $<$ 0.01), since they do not accumulate over consecutive windows. This is due to the fact that cleaning jobs are run periodically to remove temporary files, therefore decreasing disk utilization. Second, the temporal progression of events still matters, since events weigh more if they occur closer to the start of the prediction window. Third, certain types of events (e.g., read response time, read transfer size) generally have more predictive power than others (e.g., disk utilization).  

\textit{Prediction: No fail / Example 2} -- Finally, we show that the time when a predictive anomalous event occurs impacts the actual prediction. In the first example, the peak backend write response time coupled with the write response time in day 14 had the largest impact on the fail probability. In Table~\ref{nofailnice} we highlight only the occurence of peak backend write response time events for a device that does not fail. Even though these events occur multiple times, they are too distant from the start of the prediction window, thus their weights are low (i.e., 0.025 and 0.04). Additionally, no write response time events occur in the same windows. Therefore, our model is sensitive to events frequency, their exact timestamps and their co-occurence with correlated event types.

\subsection{Limitations}
As this work is only in its initial phase, it is by no means complete. We identify a few limitations to be addressed next.

\textit{Data set size} -- Given the low failure rate in our use case, ideally we should use series of KPIs collected over months or even years to increase the size of the minority class and improve prediction accuracy. For this paper, we wanted to show a proof of concept of how we can provide explanations with LSTMs, while we are continuing to collect data.

\textit{Multivariate anomalous events} -- So far, we focused on single KPI thresholds to identify anomalous events. We plan to use anomaly detection algorithms to explore multivariate time series and identify complex anomalous patterns instead. First, we expect to uncover seasonal patterns of behavior that are not obvious to an expert. Second, we believe model performance will increase and the sizes of clusters will reduce, making it easier for a domain expert to use the explanations provided. However, our primary goal was to show how we can marry LSTMs with built-in explainability to provide interpretable predictions. 

\textit{Advanced DNN models} -- We are aware that prediction accuracy can be improved by using more advanced DNN models, such as bidirectional LSTMs, RNNs with gated recurrent units~\cite{gru} or even combinations of LSTMs and CNNs~\cite{lstmcnn}. In the latter, fully convolutional blocks and LSTM units are run in parallel and their outputs are passed to a softmax classification layer. We plan to try out more advanced models going forward, while keeping the attention layer in place.

\begin{table*}
\centering
    \begin{tabular}{| l | l | l | l | l | l | l |}
\hline
\textbf{Window} & \textbf{Start timestamp} & \textbf{Duration} & \textbf{\# Events} &\textbf{Event} & \textbf{Frequency} & \textbf{Contribution} \\
\hline
1 & Day 1 10:07 & 65 min & 1& Disk utilization & 1 & 0.00 \\
\hline
2 & Day 3 10:02 & 395 min & 9& \shortstack[l]{Disk utilization\\Read transfer size} & \shortstack{5\\4} & \shortstack{0.00\\0.00} \\
\hline
3 & Day 4 2:07&195 min&2& Read transfer size & 2 & $<$0.01\\
 \hline
4 & Day 6 8:37& 15 min &2& Read response time & 2 & $<$0.01 \\
\hline
5 & Day 10 15:07  & 25 min&1& Read response time& 1 & 0.04\\
\hline
6 & Day 11 18:22 & 65 min &2& Read transfer size & 2& 0.05\\
\hline
7 & Day 13 2:47 & 135 min &5& \shortstack[l]{Read response time\\Disk utilization} & \shortstack{2\\3} & \shortstack{0.04\\0.02} \\
\hline
    \end{tabular}
\caption{Weights associated with 22 anomalous events clustered in 7 windows for a storage device predicted not to fail.}  
\label{nofail}
\end{table*}

\begin{table*}
\centering
    \begin{tabular}{| l | l | l | l | l | l | l |}
\hline
\textbf{Window} & \textbf{Start timestamp} & \textbf{Duration} & \textbf{\# Events} &\textbf{Event} & \textbf{Frequency} & \textbf{Contribution} \\
\hline
1 & Day 2 15:17 & 35 min & 5 & \shortstack[l]{Peak backend write response time\\Read response time} & \shortstack{2\\3} & \shortstack{0.05\\0.00} \\
\hline
2 & Day 5 12:02& 105 min &2& Peak backend write response time & 2 & 0.06 \\
\hline
    \end{tabular}
\caption{Highlighted weights associated with 5 peak backend write response time anomalous events clustered in 2 windows for a storage device predicted not to fail.}  
\label{nofailnice}
\end{table*}

\section{Related Work}

While there exists a lot of work around explainable models for images and text, little attention has been been given to explaining models based on temporal data, namely time or event series. On the one hand, post-hoc approaches aim at explaining a model's prediction after the event, which means they should be agnostic and applicable on any type of data. On the other hand, ante-hoc methods incorporate explainability directly into the black-box model, which implies they are tailored to the underlying model and data.

\textit{Post-hoc approaches} aim to provide local explanations for specific decisions, rather than attempting to explain the whole system behavior. One of the most representative examples for classification in recent years is LIME~\cite{lime}. The approach is simple: generate an explanation by approximating the underlying model by an interpretable one (e.g., a linear model with a only a few non-zero coefficients), learned on perturbations of the original instance. Typical perturbations can be removing words or hiding parts of an image. A similar model-agnostic approach is BETA~\cite{approximations}, which optimizes for fidelity to the black-box model and interpretability of the explanation. ~\cite{layerwise} focuses on pixel-wise decomposition of nonlinear classifiers, which allows to visualize contributions of single pixels to predictions for kernel-based classifiers. ~\cite{mining} extracts explanations from latent factor recommendation systems by training association rules on the output of a matrix factorization black-box model. All approaches have been applied on text and images, but are not built to take into consideration temporal progressions in time or event series.

\textit{Ante-hoc approaches} are interpretable by design~\cite{nphard}. Typical examples include decision trees, decision sets~\cite{decisionsets,anchors}, fuzzy inference models~\cite{pairwise} or additive models~\cite{additive}. However, none of these fit temporal data well. 

The vast majority of explainable models for time series target their classification. ~\cite{categorization} propose grammar-based decision trees to classify heterogeneous time series. 
~\cite{earlyclass,shapelets} extract interpretable features from series, expressed as local shapelets, while ~\cite{dtw} learn such shapelets via stochastic gradient learning and use them for early classification. In~\cite{irreversible}, the authors propose reversible and irreversible explainable tweaking, where given a time series and an opaque classifier, the objective is to find the minimum number of changes to the time series such that the classifier changes its decision. 

Closest to our problem is the method proposed in~\cite{iclr2018}. There, the objective is to predict a future neural event based on a sequence of previously occurred events. Current approaches are mostly concerned with time-independent sequences, in which the actual time span between events is irrelevant and the difference between events is the difference between their order positions in the sequence. The authors extract and use the information provided by the time span between events in an RNN-based model to achieve some accuracy gain over baseline models. We also opt for an RNN architecture, but we additionally incorporate attention mechanisms~\cite{attention} into the network to quantify how much an anomalous event contributed to a predicted critical incident.

\section{Conclusions}

Predictive modeling based on temporal data is key in many domains, from healthcare to IT and industries, particularly when it is concerned with critical incidents, such as failures. Providing explanations for these predictions is crucial, as it enables experts to gain trust in AI-powered models and take into consideration their outputs in the decision process. State of the art explainable models mostly focus on images and text and are not easily applicable to time or event series. We propose a deep learning approach that takes into consideration the irregularity and frequency of anomalous events extracted from time series and uses attention mechanisms to aggregate context information of these events in order to quantify how much information from each event flows into the network. A preliminary evaluation on 266081 events collected from real world storage environments shows that our approach is comparable in accuracy with traditional LSTMs, while at the same time being able to quantify the contribution of each past event recorded to a failure prediction.

\begin{quote}
\begin{small}
\bibliographystyle{aaai}
\bibliography{biblio}

\begin{thebibliography}{}

\bibitem[\protect\citeauthoryear{Bach \bgroup et al\mbox.\egroup
  }{2015}]{layerwise}
Bach, S.; Binder, A.; Montavon, G.; Klauschen, F.; Muller, K.-R.; and Samek, W.
\newblock 2015.
\newblock {On Pixel-Wise Explanations for Non-Linear Classifier Decisions by
  Layer-Wise Relevance Propagation}.
\newblock {\em PLOS One} 10(7).

\bibitem[\protect\citeauthoryear{Bai \bgroup et al\mbox.\egroup
  }{2018}]{disease}
Bai, T.; Zhang, S.; Egleston, B.~L.; and Vucetic, S.
\newblock 2018.
\newblock {Interpretable Representation Learning for Healthcare via Capturing
  Disease Progression through Time}.
\newblock {\em SIGKDD Conference on Knowledge Discovery and Data Mining, KDD}
  43--51.

\bibitem[\protect\citeauthoryear{Bascol \bgroup et al\mbox.\egroup
  }{2016}]{anomaly2}
Bascol, K.; Emonet, R.; Fromont, E.; and Odobez, J.-M.
\newblock 2016.
\newblock {Unsupervized Interpretable Pattern Discovery in Time Series using
  Autoencoders}.
\newblock {\em Workshop on Structural and Syntactic Pattern Recognition, SSPR}.

\bibitem[\protect\citeauthoryear{Bengio, Courville, and
  Vincent}{2013}]{representation}
Bengio, Y.; Courville, A.; and Vincent, P.
\newblock 2013.
\newblock {Representation Learning: A Review and New Perspectives}.
\newblock {\em IEEE Transactions on Pattern Analysis and Machine Intelligence}
  35(8):1798–--1828.

\bibitem[\protect\citeauthoryear{Brodersen \bgroup et al\mbox.\egroup
  }{2015}]{changepoint}
Brodersen, K.~H.; Gallusser, F.; Koehler, J.; Remy, N.; and Scott, S.~L.
\newblock 2015.
\newblock {Inferring Causal Impact using Bayesian Structural Time-series
  Models}.
\newblock {\em Annals of Applied Statistics} 9:247--274.

\bibitem[\protect\citeauthoryear{Caruana \bgroup et al\mbox.\egroup
  }{2015}]{pneumonia}
Caruana, R.; Lou, Y.; Gehrke, J.; Koch, P.; Sturm, M.; and Elhadad, N.
\newblock 2015.
\newblock {Intelligible Models for HealthCare: Predicting Pneumonia Risk and
  Hospital 30-day Readmission}.
\newblock {\em SIGKDD Conference on Knowledge Discovery and Data Mining, KDD}
  1721--1730.

\bibitem[\protect\citeauthoryear{Chalapathy, Menon, and
  Chawla}{2018}]{anomaly1}
Chalapathy, R.; Menon, A.~K.; and Chawla, S.
\newblock 2018.
\newblock {Anomaly Detection using One-Class Neural Networks}.
\newblock {\em SIGKDD Conference on Knowledge Discovery and Data Mining, KDD}.

\bibitem[\protect\citeauthoryear{Che \bgroup et al\mbox.\egroup
  }{2016}]{rnnmultivariate}
Che, Z.; Purushotham, S.; Cho, K.; Sontag, D.; ; and Liu, Y.
\newblock 2016.
\newblock {Recurrent Neural Networks for Multivariate Time Series With Missing
  Values}.
\newblock {\em arXiv preprint arXiv:1606.01865}.

\bibitem[\protect\citeauthoryear{Cho \bgroup et al\mbox.\egroup }{2014}]{gru}
Cho, K.; Bart Van~Merrienboer, C.~G.; Bahdanau, D.; Bougares, F.; Schwenk, H.;
  and Bengio, Y.
\newblock 2014.
\newblock {Learning Phase Representations using RNN Encoder-decoder for
  Statistical Machine Translation}.
\newblock {\em arXiv preprint arXiv:1406.1078}.

\bibitem[\protect\citeauthoryear{Choi \bgroup et al\mbox.\egroup
  }{2016}]{retain}
Choi, E.; Bahadori, M.~T.; Sun, J.; Kulas, J.; Schuetz, A.; and Stewart, W.
\newblock 2016.
\newblock {Retain: An Interpretable Predictive Model for Healthcare Using
  Reverse Time Attention Mechanism}.
\newblock {\em Advances in Neural Information Processing Systems, NIPS}
  3504--3512.

\bibitem[\protect\citeauthoryear{Ghalwash and Obradovic}{2012}]{shapelets}
Ghalwash, M.~F., and Obradovic, Z.
\newblock 2012.
\newblock {Early Classification of Multivariate Temporal Observations by
  Extraction of Interpretable Shapelets}.
\newblock {\em BMC Bioinformatics} 13(195).

\bibitem[\protect\citeauthoryear{Holzinger \bgroup et al\mbox.\egroup
  }{2017}]{nphard}
Holzinger, A.; Plass, M.; Holzinger, K.; Crisan, G.~C.; Pintea, C.-M.; and
  Palade, V.
\newblock 2017.
\newblock {A Glass-box Interactive Machine Learning Approach for Solving
  NP-hard Problems With the Human-in-the-loop}.
\newblock {\em arXiv preprint arXiv:1708.01104}.

\bibitem[\protect\citeauthoryear{Karim \bgroup et al\mbox.\egroup
  }{2017}]{lstmcnn}
Karim, F.; Majumdar, S.; Darabi, H.; and Chen, S.
\newblock 2017.
\newblock {LSTM Fully Convolutional Networks for Time Series Classification}.
\newblock {\em IEEE Access} 6:1662--1669.

\bibitem[\protect\citeauthoryear{Karlsson \bgroup et al\mbox.\egroup
  }{2018}]{irreversible}
Karlsson, I.; Rebane, J.; Papapatrou, P.; and Gionis, A.
\newblock 2018.
\newblock {Explainable Time Series Tweaking via Irreversible and Reversible
  Temporal Transformations}.
\newblock {\em International Conference on Data Mining, ICDM}.

\bibitem[\protect\citeauthoryear{Kathareios \bgroup et al\mbox.\egroup
  }{2017}]{nemo}
Kathareios, G.; Anghel, A.; Mate, A.; Clauberg, R.; and Gusat, M.
\newblock 2017.
\newblock {Catch It If You Can: Real-Time Network Anomaly Detection With Low
  False Alarm Rates}.
\newblock {\em International Conference On Machine Learning and Applications,
  ICMLA}  924--929.

\bibitem[\protect\citeauthoryear{Lakkaraju, Bach, and
  Leskovec}{2016}]{decisionsets}
Lakkaraju, H.; Bach, S.~H.; and Leskovec, J.
\newblock 2016.
\newblock {Interpretable Decision Sets: A Joint Framework for Description and
  Prediction}.
\newblock {\em SIGKDD Conference on Knowledge Discovery and Data Mining, KDD}
  1675--1684.

\bibitem[\protect\citeauthoryear{Lakkaraju \bgroup et al\mbox.\egroup
  }{2017}]{approximations}
Lakkaraju, H.; Kamar, E.; Caruana, R.; and Leskovec, J.
\newblock 2017.
\newblock {Interpretable and Explorable Approximations of Black Box Models}.
\newblock {\em arXiv preprint arXiv:1707.01154}.

\bibitem[\protect\citeauthoryear{Lee \bgroup et al\mbox.\egroup
  }{2017}]{categorization}
Lee, R.; Kochenderfer, M.~J.; Mengshoel, O.~J.; and Silbermann, J.
\newblock 2017.
\newblock {Interpretable Categorization of Heterogeneous Time Series Data}.
\newblock {\em SIGKDD Conference on Knowledge Discovery and Data Mining, KDD}.

\bibitem[\protect\citeauthoryear{Li, Du, and Bengio}{2018}]{iclr2018}
Li, Y.; Du, N.; and Bengio, S.
\newblock 2018.
\newblock {Time-Dependent Representation for Neural Event Sequence Prediction}.
\newblock {\em International Conference on Learning Representations, ICLR}.

\bibitem[\protect\citeauthoryear{lim}{2017}]{limets}
2017.
\newblock {https://github.com/emanuel-metzenthin/Lime-For-Time}.

\bibitem[\protect\citeauthoryear{Lou \bgroup et al\mbox.\egroup
  }{2013}]{pairwise}
Lou, Y.; Caruana, R.; Gehrke, J.; and Hooker, G.
\newblock 2013.
\newblock {Accurate Intelligible Models With Pairwise Interactions}.
\newblock {\em SIGKDD Conference on Knowledge Discovery and Data Mining, KDD}
  623--631.

\bibitem[\protect\citeauthoryear{Peake and Wang}{2018}]{mining}
Peake, G., and Wang, J.
\newblock 2018.
\newblock {Explanation Mining: Post Hoc Interpretability of Latent Factor
  Models for Recommendation Systems}.
\newblock {\em SIGKDD Conference on Knowledge Discovery and Data Mining, KDD}
  2060--2069.

\bibitem[\protect\citeauthoryear{Poulin \bgroup et al\mbox.\egroup
  }{2006}]{additive}
Poulin, B.; Eisner, R.; Szafron, D.; Lu, P.; Greiner, R.; Wishart, D.; Fyshe,
  A.; Pearcy, B.; MacDonell, C.; and Anvik, J.
\newblock 2006.
\newblock {Visual Explanation of Evidence in Additive Classifiers}.
\newblock  1822--1829.

\bibitem[\protect\citeauthoryear{Ribeiro, Singh, and Guestrin}{2016}]{lime}
Ribeiro, M.~T.; Singh, S.; and Guestrin, C.
\newblock 2016.
\newblock {"Why Should I Trust You?": Explaining the Predictions of Any
  Classifier}.
\newblock {\em SIGKDD Conference on Knowledge Discovery and Data Mining, KDD}
  1135--1144.

\bibitem[\protect\citeauthoryear{Ribeiro, Singh, and Guestrin}{2018}]{anchors}
Ribeiro, M.~T.; Singh, S.; and Guestrin, C.
\newblock 2018.
\newblock {Anchors: High-precision Model-agnostic Explanations}.
\newblock {\em Association for the Advancement of Artificial Intelligence,
  AAAI}  1527--1535.

\bibitem[\protect\citeauthoryear{Shah \bgroup et al\mbox.\egroup }{2016}]{dtw}
Shah, M.; Grabocka, J.; Schilling, N.; Wistuba, M.; and Schmidt-Thieme, L.
\newblock 2016.
\newblock {Learning DTW-Shapelets for Time-Series Classification}.
\newblock {\em International Conference on Data Science, CODS}.

\bibitem[\protect\citeauthoryear{Vaswani \bgroup et al\mbox.\egroup
  }{2017}]{attention}
Vaswani, A.; Shazeer, N.; Parmar, N.; Uszkoreit, J.; Jones, L.; Gomez, A.~N.;
  Kaiser, L.; and Polosukhin, I.
\newblock 2017.
\newblock {Attention Is All You Need}.
\newblock {\em Advances in Neural Information Processing Systems, NIPS}.

\bibitem[\protect\citeauthoryear{Wang and Song}{2011}]{ckmeans}
Wang, H., and Song, M.
\newblock 2011.
\newblock {Ckmeans.1d.dp: Optimal k-means Clustering in One Dimension by
  Dynamic Programming}.
\newblock {\em R Journal} 3(2):29--33.

\bibitem[\protect\citeauthoryear{Weld and Bansal}{2018}]{crafting}
Weld, D.~S., and Bansal, G.
\newblock 2018.
\newblock {The Challenge of Crafting Intelligible Intelligence}.
\newblock {\em arXiv preprint arXiv:1803.04263}.

\bibitem[\protect\citeauthoryear{Xing \bgroup et al\mbox.\egroup
  }{2011}]{earlyclass}
Xing, Z.; Pei, J.; Yu, P.~S.; and Wang, K.
\newblock 2011.
\newblock {Extracting Interpretable Features for Early Classification on Time
  Series}.
\newblock {\em EDSC-SCM}.

\bibitem[\protect\citeauthoryear{Zheng \bgroup et al\mbox.\egroup
  }{2017}]{irregularity}
Zheng, K.; Wang, W.; Gao, J.; Ngiam, K.~Y.; Ooi, B.~C.; and Yip, W. L.~J.
\newblock 2017.
\newblock {Capturing Feature-Level Irregularity in Disease Progression
  Modeling}.
\newblock {\em International Conference on Information and Knowledge
  Management, CIKM}  1579--1588.

\end{thebibliography}
\end{small}
\end{quote}

\end{document}